\documentclass[a4,a4wide]{article}

\usepackage{aaai}
\usepackage{times}
\usepackage{helvet}
\usepackage{courier}
\usepackage{latexsym}
\usepackage{emilstyle}
\usepackage{amssymb}

\begin{document}

\nocopyright

\title{A Plausibility Semantics for Abstract Argumentation Frameworks}

\author{Emil Weydert \\
\ \\
Individual and Collective Reasoning Group \\
ILIAS-CSC, University of Luxembourg}

\maketitle

\begin{abstract}\begin{quote}
\noindent We propose and investigate a simple plausibility-based extension semantics for abstract argumentation frameworks based on generic instantiations by default knowledge bases and the ranking construction paradigm for default reasoning.\footnote{This is an improved - polished and partly revised - version of my ECSQARU 2013 paper. It adds a link to structured argumentation, refines the semantic instantiation concept, and discusses attacks between inference pairs.}  
\end{quote}\end{abstract}

\section{Prologue}

The past decade has seen a flourishing of abstract argumentation theory, a coarse-grained high-level form of defeasible reasoning introduced by Dung [Dung 95]. It is characterized by a top-down perspective which ignores the logical fine structure of arguments and focuses instead on logical (conflict, support, ...) or extra-logical (preferences, ...) relations between given black box arguments so as to identify reasonable argumentative positions. One way to address the complexity of enriched argument structures carrying interacting relations, and to identify the best approaches for evaluating Dung's basic attack frameworks as well as more sophisticated argumentation systems, is to look for deeper unifying semantic foundations allowing us to improve, compare, and judge existing proposals, or to develop new ones. 

A major issue is to what extent an abstract account can adequately model concrete argumentative reasoning in the context of a sufficiently expressive, preferably defeasible logic. The instantiation of abstract frameworks by more fine-grained logic-based argument constructions and configurations is therefore an important tool for justifying or criticising abstract argumentation theories. Most of this work is however based on the first generation of nonmonotonic  formalisms, like Reiter's default logic or logic programming. While these are closer to classical logic and the original spirit of Dung's approach, it is well known that they fail to model plausible implication. In fact, they are haunted by counterintuitive behaviour and violate major desiderata for default reasoning encoded in benchmark examples and rationality postulates [Mak 94]. For instance, the only way to deal even with simple instances of specificity reasoning are opaque ad hoc prioritization mechanisms. 

The goal of the present work is therefore to supplement existing instantiation efforts with a simple ranking-based semantic model which interprets arguments and attacks by conditional knowledge bases. The well-behaved ranking construction semantics for default reasoning [Wey 96, 98, 03] can then be exploited to specify a new extension semantics for Dung frameworks which allows us to directly evaluate the plausibility of argument collections. Its occasionally unorthodox behaviour may shed a new light on basic argumentation-theoretic assumptions and concepts. 

We start with an introduction to default reasoning based on the ranking construction paradigm. After a short look at abstract argumentation theory, we show how to interpret abstract argumentation frameworks by instantiating the arguments and characterizing the attacks with suitable sets of conditionals describing constraints over ranking models. Based on the concept of generic instantiations, i.e.~using minimal assumptions, and plausibility maximization, we then specify a natural ranking-based extension semantics. We conclude with a simple algorithm, some instructive examples, and the discussion of several important properties.

\section{Ranking-based default reasoning}

We assume a basic language $L$ closed under the usual propositional connectives, together with a classical satisfaction relation $\models$ inducing a monotonic entailment relation $\fs\ \subseteq 2^L\times L$. The model sets of $(L,\models)$ are denoted by $[\![\ph]\!] = \{m\mid m\models\ph\}$, resp.~$[\![\Sigma]\!] = \cap_{\ph\in\Sigma}[\![\ph]\!]$ for $\Sigma\subseteq L$. $\Bx_L$ is the boolean proposition algebra over $\BB_L = \{[\![\ph]\!]\mid\ph\in L\}$. Let $Cn(\Sigma) = \{\psi\mid\Sigma\fs\psi\}$. 

Default inference is an important instance of nonmonotonic reasoning concerned with drawing reasonable but potentially defeasible conclusions from knowledge bases of the form $\Sigma\cup\Delta$, where $\Sigma\subseteq L$ is a set of assumptions or facts, e.g.~encoding knowledge about a specific state of affairs in the domain language $L$, and $\Delta\subseteq L(\LI,\leadsto)$ is a collection of conditionals expressing strict or exception-tolerant implicational information over $L$, which is used to guide defeasible inference. $L(\LI,\leadsto) = \{\ph\LI\psi\mid\ph,\psi\in L\}\cup\{\ph\leadsto\psi\mid\ph,\psi\in L\}$ is the corresponding flat conditional language on top of $L$. In the following we will focus on finite $\Sigma$ and $\Delta$. $\Delta^{\I} = \{\ph\I\psi\mid \ph\LI\psi,\ph\leadsto\psi\}$ collects the material implications corresponding to the conditionals in $\Delta$.

The strict implication $\ph\LI\psi$ states that $\ph$ necessarily implies $\psi$, forcing us to accept $\psi$ given $\ph$. The default implication $\ph\leadsto\psi$ tells us that $\ph$ plausibly/normally implies $\psi$, and only recommends the acceptance of $\psi$ given $\ph$. The actual impact of a default depends of course on the context $\Sigma\cup\Delta$ and the chosen nonmonotonic inference concept $\fp$, which will be discussed later. 

We can distinguish two perspectives in default reasoning: the autoepistemic/context-based one, and the plausibilistic/quasi-probabilistic one. The former is exemplified by Reiter's default logic, where defaults are usually modeled by normal default rules of the form $\ph:\psi/\psi$ (if $\ph$, and it is consistent that $\psi$, then $\psi$). A characteristic feature is that the conclusions are obtained by intersecting suitable equilibrium sets, known as extensions.

The alternative is to use default conditionals interpreted by some preferential or valuational semantics, e.g. System Z [Pea 90, Leh 92], or probabilistic ME-based accounts [GMP 93] (ME = maximum-entropy). For historical reasons and technical convenience (closeness to classical logic), the first approach has received most attention, especially in the context of argumentation. However, this ignores the fact that the conditional semantic paradigm has a much better record when it comes to the natural handling of benchmark examples and the satisfaction of rationality postulates [Mak 94]. It therefore seems promising to investigate whether such semantic-based accounts can also help to instantiate and evaluate abstract argumentation frameworks.  

Our default conditional semantics for interpreting argumentation frameworks is based on the simplest plausibility measure concept able to reasonably handle independence and conditionalization, namely Spohn's ranking functions [Spo 88, 12], or more generally, ranking measures [Wey 94]. These are quasi-probabilistic belief valuations expressing the degree of surprise or implausibility of propositions. Integer-valued ranking functions were originally introduced by Spohn to model the iterated revision of graded plain belief. We will consider $[0,\infty]_{real}$-valued ranking measures\footnote{Although for us, rational values would actually be sufficient.}, where $\infty$ expresses doxastic impossibility. 

\bd[Real-valued ranking measures] \ \\
A map $R:\Bx_L\I ([0,\infty],0,\infty,+,\leq)$ is called a real-valued ranking measure iff $R([\![\T]\!]) = 0$, $R([\![\F]\!]) = R(\emptyset) = \infty$, and for all $A,B\in\BB_L$, $R(A\cup B) = \min_{\leq}\{R(A),R(B)\}$. $R(.|.)$ is the associated conditional ranking measure defined by $R(B|A) = R(A\cap B)-R(A)$ if $R(A)\neq\infty$, else $R(B|A) = \infty$. $R_0$ is the uniform ranking measure, i.e.~$R_0(A) = 0$ for $A\neq\emptyset$. If $\Bx = \Bx_L$, we will use the abbreviation $R(\ph) := R([\![\ph]\!])$. 
\ed

\noindent For instance, the order of magnitude reading interprets ranking measure values $R(A)$ as exponents of infinitesimal probabilities $P(A) = p_A\eps^{R(A)}$, which explains the parallels with probability theory. The monotonic semantics of our conditionals $\LI,\leadsto$ is based on the satisfaction relation $\models_{rk}$. The corresponding truth conditions are
\bi
\item \ $R\models_{rk}\ph\LI\psi$ \ iff \ $R(\ph\wedge\neg\psi) = \infty$.
\item \ $R\models_{rk}\ph\leadsto\psi$ \ iff \ $R(\ph\wedge\psi)+1\leq R(\ph\wedge\neg\psi)$. 
\ei
That is, we assume that a strict implication $\ph\LI\psi$ states that $\ph\wedge\neg\psi$ is doxastically impossible. 

Note that we may replace $\ph\LI\psi$ by $\ph\wedge\neg\psi\leadsto\F$, i.e.~it would be actually enough to consider $L(\leadsto)$. We use $\leq$ with a threshold because this provides more discriminatory power and also guarantees the existence of minima for relevant ranking construction procedures. 
The exchangeability of arbitrary $r,r'\neq 0,\infty$ by automorphisms allows us to focus, by convention, on the threshold $1$. For $\Delta\cup\{\delta\}\subseteq L(\LI,\leadsto)$, we set 
\bc
	$[\![\Delta]\!]_{rk} = \{R \mid R\models_{rk}\Delta\}$, \ $\Delta\fs_{rk}\delta$ iff $[\![\Delta]\!]_{rk}\subseteq [\![\delta]\!]_{rk}$. 
\ec
$\fs_{rk}$ is monotonic and verifies the axioms and rules of preferential conditional logic and disjunctive rationality (threshold semantics: no rational monotony) for $\leadsto$ [KLM 90]. 

But it is important to understand that the central concept in default reasoning is not some monotonic conditional logic for $L(\LI,\leadsto)$, but a nonmonotonic meta-level inference relation $\fp$ over $L\cup L(\LI,\leadsto)$ specifying which conclusions $\psi\in L$ can be plausibly inferred from finite $\Sigma\cup\Delta \subseteq L\cup L(\LI,\leadsto)$. We write $\Sigma\cup\Delta\fp\psi$, or alternatively $\Sigma\fp_{\Delta}\psi$, and set $C_{\Delta}^{\fp}(\Sigma) = \{\psi\mid \Sigma\fp_{\Delta}\psi\}$.

The ranking semantics for plausibilistic default reasoning is based on nonmonotonic ranking choice operators $\Ix$ which map each finite $\Delta\subseteq L(\LI,\leadsto)$ to a collection $\Ix(\Delta)\subseteq [\![\Delta]\!]_{rk}$ of preferred ranking models of $\Delta$. A corresponding ranking-based default inference notion $\fp^{\Ix}$ can then be specified by  
\bc 
	$\Sigma\fp^{\Ix}_{\Delta}\psi$ \ iff \ for all $R\in\Ix(\Delta)$, $R\models_{rk}\wedge\Sigma\leadsto\psi$.
\ec
Similarly, we can also define a monotonic inference concept characterizing the strict consequences. 
\bc
	$\Sigma\fs^{\Ix}_{\Delta}\psi$ \ iff \ for all $R\in\Ix(\Delta)$, $R\models_{rk}\wedge\Sigma\LI\psi$.
\ec
If $\Ix(\Delta) = [\![\Delta]\!]_{rk}$, $\fp^{\Ix}_{\Delta}$ is, modulo cosmetic details, equivalent to preferential entailment (System P) [KLM 90]. If $\leq_{pt}$ describes the pointwise comparison of ranking measures, i.e.~$R\leq_{pt}R'$ iff for all $A\in\BB_L$ $R(A)\leq R'(A)$, then $\Ix(\Delta) = \{Min_{\leq_{pt}}[\![\Delta]\!]_{rk}\}$ essentially characterizes System Z [Pea 90]. Because these approaches fail to adequately deal with inheritance to exceptional subclasses, we introduced and developed the construction paradigm for default reasoning [Wey 96, 98, 03], which is a powerful strategy for specifying reasonable $\Ix$ based on Spohn's Jeffrey-conditionalization for ranking measures. The resulting default inference notions are well-behaved and show nice inheritance features. The essential idea is that defaults do not only specify ranking constraints, but also admissible construction steps to generate them. In particular, for each default $\ph\leadsto\psi$, we are allowed to uniformly shift upwards (make less plausible/increase the ranks of) the $\ph\wedge\neg\psi$-worlds, which amounts to strengthen belief in the corresponding material implication $\ph\I\psi$. If $W$ is finite, this is analogous to specifying the rank of a world by adding a weight $\geq 0$ for each default it violates. More formally, we define a shifting transformation $R\I R+r[\rho]$ such that for each ranking measure $R$, $\chi,\rho\in L$, and $r\in [0,\infty]$, we set 
\bc
	$(R+r[\rho])(\chi) = min\{R(\chi\wedge\rho)+r,R(\chi\wedge\neg\rho)\}$.
\ec
This corresponds to uniformly shifting $\rho$ by $r$.

\bd[Constructibility] \ \\ Let $\Delta = \{\ph_i\leadsto/\LI\psi_i\mid i\leq n\}\subseteq L(\LI,\leadsto)$. A ranking measure $R'$ is said to be constructible from $R$ over $\Delta$, written $R'\in Constr(\Delta,R)$, iff there are $r_i\in [0,\infty]$ s.t.~$R' = R+\Sigma_{i\leq n}r_i[\ph_i\wedge\neg\psi_i]$.\footnote{Similar ideas can be found in [BSS 00, KI 01].}
\ed

\noindent For instance, we obtain a well-behaved robust default inference relation, System J [Wey 96], just by setting $\Ix_J(\Delta) = Constr(\Delta,R_0)\cap [\![\Delta]\!]_{rk}$. To implement shifting minimization, we may strengthen System J by allowing proper shifting ($r_i>0$) only  if the targeted ranking constraint interpreting a default $\ph_i\leadsto\psi_i$ is realized as an equality constraint $R(\ph_i\wedge\psi_i)+1 = R(\ph_i\wedge\neg\psi_i)$. Otherwise, the shifting wouldn't seem to be justified in the first place. 

\bd[Justifiable constructibility]\ \\
$R$ is called a justifiably constructible model of $\Delta$, written $R\in\Ix_{jj}(\Delta)$ iff $R\models_{rk}\Delta$, $R = R_0+\Sigma_{i\leq n}r_i[\ph_i\wedge\neg\psi_i]$, and for each $r_j > 0$, $R(\ph_j\wedge\psi_j)+1 = R(\ph_j\wedge\neg\psi_j)$.  
\ed

\noindent It follows from a standard property of entropy maximization (ME) that the order-of-magnitude translation of ME, in the context of a nonstandard model of the reals with infinitesimals [GMP 93, Wey 95], to the ranking level always produces a canonical justifiably constructible ranking model $R_{me}$. We set $\Ix_{me}(\Delta) = \{R^{\Delta}_{me}\}$. Hence, if $\Delta\nfs_{rk}\F$, $R^{\Delta}_{me}\in\Ix_{jj}(\Delta)\neq\emptyset$. If $\Ix_{jj}(\Delta)$ is a singleton, we have therefore $\fp^{jj} =\ \fp^{me}$. This holds for instance for minimal core default sets $\Delta$ [GMP 93], where no doxastically possible $\ph_i\wedge\neg\psi_i$, i.e.~$\Delta\nfs_{rk}\ph_i\wedge\neg\psi_i\leadsto\F$, is covered by other $\ph_j\wedge\neg\psi_j$. However, because of its fine-grained quantitative character, $\fp^{me}$ is actually representation-dependent, i.e.~the solution depends on how we describe a problem in $L$, it is not invariant under boolean automorphisms of $\Bx_L$. Fortunately, there are two other natural representation-independent ways to pick up a canonical justifiably constructible model. 
\bi
\item System JZ is based on on a natural canonical hierarchical ranking construction in the tradition of System Z and ensures justifiable constructibility [Wey 98, 03]. It constitutes a uniform way to implement the minimal information philosophy at the ranking level.  
\item System JJR is based on the fusion of the justifiably constructible ranking models of $\Delta$, i.e.~$\Ix_{jjr}(\Delta) = \{R^{\Delta}_{jjr}\}$, where for all $A\in\BB_L$, $R^{\Delta}_{jjr}(A) = Min_{\leq_{pt}}\Ix_{jj}(\Delta)$. $\fp^{jjr}$ may be of particular interest because its canonical ranking model is at least as plausible as every justifiably constructible one.
\ei
Note that for non-canonical $\Ix_{jj}(\Delta)$, it is possible that $R^{\Delta}_{jjr}\not\in\Ix_{jj}(\Delta)$. We have $\fp^{jj}\ \subset\ \fp^{me},\fp^{jz},\fp^{jjr}$. Fortunately, for the generic default sets we will use to interpret abstract argumentation frameworks, all four turn out to be equivalent. To conclude this section, let us consider a simple example with a single JJ-model. \\

\noindent {\bf Big birds example:} \\
\noindent Non-flying birds are not inferred to be small. 
\bc
	$\{B,\neg F\}\cup\{B\leadsto S, B\leadsto F, \neg S\leadsto \neg F\}\nfp^{jj} S$  
\ec
\noindent The canonical JJ/ME/JZ/JJR-model is then \\
\noindent $R = R_0+1[\neg F]+2[\neg S\wedge F]$. But $R\not\models_{rk} B\wedge\neg F\leadsto S$ \\
\noindent because $R(B\wedge\neg F\wedge S) = R(B\wedge\neg F\wedge\neg S) = 1$

\section{Abstract argumentation}

The idea of abstract argumentation theory, launched by Dung [Dun 95], has been to replace the traditional bottom-up strategy, which models and exploits the logical fine structure of arguments, by a top-down perspective, where arguments become black boxes evaluated only based on knowledge about specific logical or extra-logical relationships connecting them. It is interesting to see that such a coarse-grained relational analysis often seems sufficient to determine which collections of instantiated arguments are reasonable. In addition to possible conceptual and computational gains, the abstract viewpoint offers furthermore a powerful methodological tool for general argumentation-theoretic investigations.   

An abstract argumentation framework in the original sense of Dung is a structure of the form $\Ax = (\AAA,\rhd)$, where $\AAA$ is a collection of abstract entities representing arguments, and $\rhd$ is a possibly asymmetric binary attack relation modeling conflicts between arguments. To grasp the expressive complexity of real-world argumentation, several authors have extended this basic account to include further inferential or cognitive relations, like support links, preferences, valuations, or collective attacks. Our general definition\footnote{A bit of an overkill for this paper, but we couldn't resist.} [Wey 11] for the first-order context is as follows. 

\bd[Hyperframeworks] A general abstract argumentation framework, or hyperframework (HF), is just a structure of the form $\Ax = (\AAA,(\Rx_i)_{i\in I},(\Px_j)_{j\in J})$, where $\AAA$ is the domain of arguments, the $\Rx_i$ are conflictual, and the $\Px_i$ non-conflictual relations over $\AAA$. $B\subseteq\AAA$ is said to be conflict-free iff it does not instantiate a conflictual relation.   
\ed

\noindent For instance, standard Dung frameworks $(\AAA,\rhd)$ carry one conflictual and no non-conflictual relations. The general inferential task in abstract argumentation is to identify reasonable evaluations of the arguments described by $\Ax$, e.g.~to find out which sets of arguments describe acceptable argumentative positions. These are called extensions. In Dung's scenario, the extensions are $E\subseteq\AAA$ obeying suitable acceptability conditions in the context of $\Ax$, the minimal requirement being the absence of internal conflicts. For instance, $E$ is admissible iff it is conflict-free and each attacker of an $a\in E$ is attacked by some $b\in E$. $E$ is grounded/preferred iff it is minimally/maximally admissible, it is stage iff $E\cup\rhd''E$ is maximal, semi-stable if it is also admissible, and stable iff $\AAA-E = \rhd''E$. Here $\rhd''E$ is the relational image of $E$, i.e.~the set of $a\in\AAA$ attacked by some $b\in E$. In concrete decision contexts, we may however also want to exploit finer-grained assessments of arguments, like prioritizations or classifications. This suggests a more general semantic perspective [Wey 11]. 

\bd[Hyperextensions] A hyperframework semantics is a map $\Ex$ associating with each hyperframework $\Ax = (\AAA,(\Rx_i)_{i\in I},(\Px_j)_{j\in J})$ of a given signature a collection $\Ex(\Ax)$ of distinguished evaluation structures expanding $\Ax$, of the form $(\Ax,In^{\Ax},(\Fx_h)_{h\in H})$. $In^{\Ax}$ is here a conflict-free subset of $\AAA$. The elements of $\Ex(\Ax)$ are called hyperextensions of $\Ax$. 
\ed

\noindent $In^{\Ax}$ plays here the role of a classical extension, whereas the $\Fx_h$ ($h\in H$) express more sophisticated structures over arguments, e.g.~a posteriori plausibility orderings, value predicates, or completions of framework relations considered partial. If $H = \emptyset$, we are back to Dung.

\section{Concretizing arguments}

Ideally, abstract argumentation frameworks should be reconstructible as actual abstractions of logic-based argumentation scenarios. Such an anchoring seems required to develop, evaluate, and apply the abstract models in a suitable way. In a first step, this amounts to instantiate the abstract arguments from the framework domain by logical entities representing concrete arguments, and to interpret the abstract framework structure by specific inferential or evaluational relationships fitting the conceptual intentions the abstract level tries to capture. In what follows we will sketch a natural hierarchy of instantiation layers, passing from more concrete, deep instantiations, to more abstract, shallow ones, with a focus on the intermediate level. \\

\noindent {\bf Structured instantiations:} \\
\noindent We start with logic-based structured argumentation over a defeasible conditional logic $\Lx_{\delta} = (L\cup L(\LI,\leadsto),\fs^{\delta},\fp^{\delta})$, with $(L,\fs)$ as a classical Tarskian background logic. For the moment, we do not impose any further a priori conditions on $\Lx_{\delta}$. But eventually we will turn to specific ranking-based default formalisms. In the context of $\Lx_{\delta}$, a concrete defeasible argument $a$ for a claim $\psi_a\in L$, exploiting some given general knowledge base $\Sigma\cup\Delta$, is modeled by a finite rooted defeasible inference tree $\Tx_a$ whose nodes $s$ are tagged by local claims $\eta_s\in L\cup L(\LI,\leadsto)$ such that
\bi
\item the root node is tagged by $\psi_a$, 
\item the leaf nodes are tagged by $\eta_s\in\Sigma_a\cup\Delta_a\cup\Lambda$, where $\Lambda = \{\T\}\cup\{\ph\LI\ph,\ph\leadsto\ph\mid\ph\in L\}$ (basic tautologies),
\item the non-leaf nodes are tagged by $\eta_s\in L$ s.t.~$\Gamma_s\fp^{\delta}\eta_s$ where $\Gamma_s$ is the set of claims from the children of $s$. 
\ei
$\Sigma_a\cup\Delta_a$ is the contingent premise set of $a$, the premises being the claims of the leaf nodes. Within concrete arguments, the local justification steps, e.g.~from $\Gamma_s$ to $\eta_s$, are typically assumed to be elementary, like instances of modus ponens. To handle reasoning by cases, which holds for plausible implication, we may also apply the {\it disjunctive modus ponens} for $\LI$ and $\leadsto$, e.g. 
\bc
	$\Gamma_s = \{\ph_1\vee\ldots\vee\ph_n,\ph_1\leadsto\psi_1,\ldots,\ph_n\LI\psi_n\}$ \\
 	\ \\
	$\fp^{\delta} \psi_1\vee\ldots\vee\psi_n (= \eta_s)$.
\ec
If $\Gamma_s\subseteq L\cup L(\LI)$, we can replace $\fp^{\delta}$ by $\fs^{\delta}$ and obtain a strict inference step. For our purposes we may ignore the exact nature of the justification steps. Note that the correctness of local inference steps does not entail the global correctness of the argument $a$. Consider for instance $\Sigma_a\cup\Delta_a = \{\ph\}\cup\{\ph\leadsto\psi,\psi\leadsto\neg\ph\}$, which is consistent w.r.t.~$\fp^{\delta} = \ \fp^{\Ix}$.  
\bc
	 $\{\ph\}\cup\{\ph\leadsto\psi\} \fp^{\delta} \psi$ and $\{\psi\}\cup\{\psi\leadsto\neg\ph\} \fp^{\delta} \neg\ph$,
\ec
but $\Sigma_a\cup\Delta_a\nfp^{\delta}\neg\ph$. This example looks odd because accepting the whole argument would require the acceptance of all its claims, which is blocked by $\ph,\neg\ph\fs\F$. In fact, a natural requirement for an acceptable argument $a$ would be  
\bc
	{\it Material consistency}: \ $\Sigma_a\cup\Delta_a^{\I}\nfs\F$. 
\ec
This means that the factual premises and the material implications corresponding to the conditional premises are classically consistent. Note that this condition is strictly stronger than $\Sigma_a\cup\Delta_a\nfp^{\delta}\F$ because we typically have $\{\T\leadsto\ph,\neg\ph\}\nfp^{\delta}\F$ whereas $\{\T\I\ph,\neg\ph\}\fs\F$. However, in practice, without omniscience w.r.t.~propositional logic, it may not be clear whether these global conditions are actually satisfied. Real arguments may well be inconsistent in the strong sense. 

In structured argumentation, an argument tree has two functions: first, to describe and offer a prima facie justification for a claim, and secondly, to specify target points where other arguments may attack. It is essentially a computational tool which is intended to help identifying - or even defining - inferential relationships within a suitable defeasible conditional logic $\Lx_{\delta}$, and to help specifying attack relations to determine reasonable argumentative positions. 

But what can we say about the semantic content of an argument represented by such a tree? What is an agent committed to if he accepts or believes a given argument, or a whole collection of arguments? Which tree attributes have to be known to specify this content? What is the meaning of attacks between arguments? \\

\noindent {\bf Conditional instantiations:} \\
Our basic idea is that, whatever the requirements for argumentation trees in the context of $\Lx_{\delta}$, and whatever the content of an argument $a$ represented by such a tree $\Tx_a$, it should only depend on the collection of local claims $\{\eta_s\mid s$ node of $\Tx_a\}$, and more specifically, on the choice of the main claim $\psi_a$, the premise claims $\Sigma_a\cup\Delta_a$, and the intermediate claims $\Psi_a$. In fact, because the acceptance of a structured argument includes the acceptance of all its subarguments, we have to consider the main claims of the subarguments as well. So we can assume that the content of $\Tx_a$ is fixed by the triple $(\Sigma_a\cup\Delta_a,\Psi_a,\psi_a)$. An agent accepting $a$ obviously has to be committed to all the elements of the base $\Sigma_a\cup\Delta_a\cup\Psi_a\cup\{\psi_a\}$. 

To be fully acceptable w.r.t.~$\Lx_{\delta}$, the structured argument also has to be globally correct in the sense that all its local claims are actually defeasibly entailed by $\Sigma_a\cup\Delta_a$. In particular, $\Sigma_a\cup\Delta_a\fp^{\delta}\psi$ for each $\psi\in\Psi_a\cup\{\psi_a\}$. This requirement should also hold for each subargument $b$ of $a$. But note that, because of defeasibility, this does not exclude that the premises $\Sigma_b\cup\Delta_b$ of a subargument $b$ could implicitly infer the negation of a local claim $\psi_x$ external to $b$, as long as this conflicting inference is eventually overridden by the full premise set $\Sigma_a\cup\Delta_a$. It follows that the strengthening of a subargument by choosing a stronger claim could undermine global correctness. But if the intermediate claims are always inferred and therefore implicitly present, we may actually drop $\Psi_a$ and just consider for each globally correct argument $a$ the finite inference pair $(\Sigma_a\cup\Delta_a,\psi_a)$. 

Given a pure Dung framework $\Ax = (\AAA,\rhd)$ and the defeasible conditional logic $\Lx_{\delta}$, a structured instantiation $I_{str}$ of $\Ax$ maps each $a\in\AAA$ to a globally correct argument tree $\Tx_a$ over $\Lx_{\delta}$. On the most general level, we do not want to impose a priori further restrictions beyond inferential correctness. In practice one may however well decide to focus on specific argument trees, e.g.~those using specific justification steps. Each $I_{str}(a)$ specifies a correct inference pair $I_{log}(a) = (\Sigma_a\cup\Delta_a,\psi_a)$, which we call a conditional logical instantiation of $a$ over $\Lx_{\delta}$. $I_{log}$ specifies the intended logical content of an argument on the syntactic level. Note that it depends on the tree concept whether we can obtain all the correct inference pairs. 

In monotonic argumentation, the consistency and minimality of the premise sets are standard assumptions. But within defeasible argumentation, a more liberal perspective may be preferable. For instance, on the structured level, we want to allow arguments claiming $\F$. The reductio ad absurdum principle then offers a possibility to attack arguments from within. Consequently, we also have to accept instantiating inference pairs whose conclusion is $\F$. On the other hand, material consistency, the existence of models of $\Sigma_a$ which do not violate any conditional in $\Delta_a$, is a natural requirement in the context of argumentation theory. But we can replace it by a qualified version, restricted to those instances where $\Sigma_a\cup\Delta_a$ is actually consistent. 

What about minimality? First, it may obviously fail for inference pairs obtained by flattening argument trees. Of course, we could consider an additional minimization step where we replace each $(\Sigma_a\cup\Delta_a,\psi_a)$ by all those $(\Phi,\psi_a)$ with $\Phi\subseteq\Sigma_a\cup\Delta_a$ and which are minimal s.t.~$\Phi\fp^{\delta}\psi_a$. Although this may be computationally costly, it could be theoretically appealing. However, minimality could also be questioned because by adding premises, a conclusion may successively get accepted, rejected, and accepted again, letting the character of the inferential support change between different levels of specifity, which calls for a discrimination between the corresponding inference pairs. Proponents of minimality object that these types of support could, perhaps, also be reproduced by suitable minimal $(\Phi,\psi_a)$. However, this assumption is not sustainable for ranking-based semantics for argumentation, because here the results may change if we restrict ourselves to minimal premise sets. In fact, shrinking $\Sigma_a\cup\Delta_a$ to $\Phi$ may actually increase the set of possible attacks. In particular, we could have attacks on all the minimal $\Phi\fp^{\delta}\psi_a$, but none on $\Sigma_a\cup\Delta_a\fp^{\delta}\psi_a$. Hence premise minimality may fail. \\ 

\noindent {\bf Shallow instantiations:} \\
\noindent Let us recall our task: exploiting a ranking semantics for default reasoning to provide a plausibilistic semantics for abstract argumentation. But inference pairs, which populate the conditional logical instantiation level, are still rather complex and opaque objects. To model argumentation frameworks and their semantics, we would here have to deal with sets of sets of conditionals, whose inferential interactions may furthermore be hard to assess. We therefore prefer to start with simpler entities and to seek more abstraction.

Consider the main goal of an agent: to extract from argument configurations suitable beliefs, expressed in the domain language $L$, whose plausibility is semantically modeled by ranking measures over $\BB_L$. Given an inference pair $(\Sigma_a\cup\Delta_a,\psi_a)$ representing the full conditional logical content of an argument $a$, in addition to the main claim $\psi_a$, there are three relevant collections of formulas: $\Sigma_a, C_{\fs}(\Delta_a\cup, \Sigma_a), C_{\fp}(\Delta_a\cup\Sigma_a)$ which represent resp.~the premises, the strict, and the defeasible consequences. If the language is finitary, this gives us four $L$-formulas representing the relevant propositional $L$-content.
\bi
\item $\ph_a = \wedge C_{\fs}(\Sigma_a)$ (premise content).  
\item $\theta_a = \wedge C_{\fs}(\Delta_a\cup\Sigma_a)$ (strict content).
\item $\delta_a = \wedge C_{\fp}(\Delta_a\cup\Sigma_a)$ (defeasible content).
\item $\psi_a$ (main claim).
\ei
We have $\delta_a \fs \theta_a \fs \ph_a$, and $\delta_a\fs\psi_a$ by inferential correctness. $\delta_a$ specifies the strongest possible claim based on the information made available by the argument. For our semantic modeling purposes, we will assume that $\psi_a = \delta_a$. If we abstract away from the representational details, we arrive at our central concept: the {\it shallow semantic instantiation} of $a$ extracted from the conditional logical instantiation $I_{log}(a)$. 
\bc
	$I_{sem}(a) = ([\![\ph_a]\!], [\![\theta_a]\!], [\![\psi_a]\!])$.
\ec
\noindent In the following, we will sloppily denote $I_{sem}(a)$ by $(\ph_a,\theta_a,\psi_a)$. One should emphasize that these propositional semantic profiles are not intended to grasp the full nature of arguments, but only to reflect certain characteristics exploitable by suitable argumentation semantics. We observe that each proposition triple $(\ph,\theta,\psi)$ with $\psi\fs\theta\fs\ph$ can become a shallow instantiation. In fact, if $I_{log}(a) = (\{\ph,\ph\LI\theta,\ph\leadsto\psi\},\psi)$, for standard $\fp^{\delta}$, we obtain $I_{sem}(a) = (\ph,\theta,\psi)$. In terms of ranking constraints, this gives us $R(\ph\wedge\neg\theta) = \infty$ and $R(\ph\wedge\psi)+1\leq R(\ph\wedge\neg\psi)$.

\section{Concretizing attacks}

One argument attacks another argument if accepting the first interferes with the inferential structure or goal of the second one. To avoid a counterattack, the premises of the attacked argument should also not affect the inferential success of the attacker, otherwise the presupposition of the attack could be undermined. In the following we will investigate attack relations between conditional logical resp.~shallow semantic instantiations of abstract arguments. We start with the former. Let $I_{log}(a) = (\Sigma_a\cup\Delta_a,\psi_a)$ and $I_{log}(b) = (\Sigma_b\cup\Delta_b,\psi_b)$ be two correct inference pairs for $\Lx_{\delta}$. We distinguish two scenarios: unilateral and mutual attack. The idea is to say that $(\Sigma_a\cup\Delta_a,\psi_a)$ unilaterally attacks $(\Sigma_b\cup\Delta_b,\psi_b)$ iff the premises of both arguments together with $\psi_a$ enforce the strict rejection of $\psi_b$, i.e.
\bc
	$\Sigma_b\cup\Delta_b\cup\Sigma_a\cup\Delta_a\cup\{\psi_a\}\fs^{\delta}\neg\psi_b$,
\ec
whereas the defeasible inference of $\psi_a$ from the premises is preserved, i.e.
\bc
	$\Sigma_a\cup\Delta_a\cup\Sigma_b\cup\Delta_b\fp^{\delta}\psi_a$.
\ec	
On the other hand, $(\Sigma_a\cup\Delta_a,\psi_a)$ and $(\Sigma_b\cup\Delta_b,\psi_b)$ attack each other iff they strictly reject each other's claims, i.e.
\bc
	$\Sigma_b\cup\Delta_b\cup\Sigma_a\cup\Delta_a\cup\{\psi_a\}\fs^{\delta}\neg\psi_b$, \ and \\
	\ \\
	$\Sigma_a\cup\Delta_a\cup\Sigma_b\cup\Delta_b\cup\{\psi_b\}\fs^{\delta}\neg\psi_a$. 
\ec 
This holds for instance if their premise sets, resp.~their claims, are classically inconsistent. This definition provides one of the strongest possible natural attack relations for inference pairs. Note that we have a self-attack iff the premise set is inconsistent, i.e.~$\Sigma_a\cup\Delta_a\fs^{\delta}\F$. To exploit the powerful semantics of ranking-based default reasoning, in what follows we will assume that $\fp^{\delta} =\ \fp^{\Ix}$, where $\Ix$ is a ranking choice function. 
   
How can we exploit the above approach to define attacks between shallow instantiations, e.g.~$I_{sem}(a) = (\ph_a,\theta_a,\psi_a)$ and $I_{sem}(b) = (\ph_b,\theta_b,\psi_b)$? The corresponding inference pairs are $I_{log}(a) = (\{\ph_a,\ph_a\LI\theta_a,\ph_a\leadsto\psi_a\},\psi_a)$ and $I_{log}(b) = (\{\ph_b,\ph_b\LI\theta_b,\ph_b\leadsto\psi_b\},\psi_b)$. For an unilateral attack from $I_{log}(a)$ on $I_{log}(b)$, we must have
\bc
	$\Ix(\Delta_a\cup\Delta_b)\models_{rk}\ph_a\wedge\ph_b\wedge\psi_a\LI\neg\psi_b$, \ and \\
	\ \\
	$\Ix(\Delta_a\cup\Delta_b)\models_{rk}\ph_a\wedge\ph_b\leadsto\psi_a$.
\ec
This is, for instance, automatically realized if $\psi_a\fs\neg\psi_b$, $\ph_a\fs\ph_b$, and we have logical independence elsewhere. For a bilateral attack, we may just drop the condition $\ph_a\fs\ph_b$. However, we do not have to presuppose that all the attacks result from the logical structure induced by the instantiation. In fact, in addition to the instantiation-intrinsic attack relationships, there could be further attack links derived from a separate conditional knowledge base reflecting other known attacks.   

From a given Dung framework $\Ax = (\AAA,\rhd)$ and a shallow instantiation $I = I_{sem}$, if we adopt the ranking semantic perspective and the above attack philosophy, we can induce a collection of conditionals specifying ranking constraints. For any $a\in\AAA$, the shallow inference pair supplies $\ph_a\LI\theta_a$ (alternatively, $\ph_a\wedge\neg\theta_a\leadsto\F$) and $\ph_a\leadsto\psi_a$. For every attack $a\rhd b$, we get at least $\psi_a\wedge\psi_b\leadsto\F$. Note that this is a consequence of choosing maximal claims at the instantiation level. For each unilateral attack $a\rhd b$ we must add $\ph_a\wedge\ph_b\leadsto\psi_a$ to preserve the inferential impact of $a$ in the context of $b$. The resulting default base is 
\bc
	$\Delta^{\Ax,I} = \{\ph_a\leadsto\psi_a,\ph_a\LI\theta_a\mid a\in\AAA\}$ \\
        $\cup\ \{\psi_a\wedge\psi_b\leadsto\F\mid a\rhd b $ or $ b\rhd a\}$ \\
        $\cup\ \{\ph_a\wedge\ph_b\leadsto\psi_a\mid a\rhd b, b\not\!\rhd\ a\}$.
\ec
We observe that for each 1-loop, we get $\psi_a\leadsto\F$ and $\ph_a\leadsto\psi_a$, hence $\Delta^{\Ax,I}\fs^{\Ix}\neg\ph_a$. The doxastic impossibility of $\ph_a$ illustrates the paradoxical character of self-attacking arguments. The belief states compatible with an instantiated framework $\Ax,I$ are here represented by the ranking models of $\Delta^{\Ax,I}$.

Conversely, we can identify for each instantiation $I$ of $\AAA$ and each collection of ranking measures $\Rx\models_{rk}\Delta^{\AAA,I} = \{\ph_a\leadsto\psi_a,\ph_a\LI\theta_a\mid a\in\AAA\}$ the attacks  supported by all the $R\in\Rx$. Let $\rhd^{\Rx}_I$ be the resulting attack relation, that is, for each $a,b\in\AAA$
\bc
	$a\rhd^{\Rx}_I b$ \ iff \ for all $R\in\Rx$, $R\models_{rk}\psi_a\wedge\psi_b\leadsto\F$ and \\
	\ \\
       ($R\models_{rk}\ph_a\wedge\ph_b\leadsto\psi_a$ or $R\not\models_{rk}\ph_a\wedge\ph_b\leadsto\psi_b$).
\ec
The second disjunct is the result of an easy simplification. If $a$ or $b$ are self-reflective, we have $a\rhd^{\Rx}_I b$ because conditionals always hold if the premises are doxastically impossible. 
Because in this paper we will mainly consider canonical ranking choice functions, we are going to focus on $\Rx = \{R\}$, setting $\rhd^R_I = \rhd^{\Rx}_I$. 

\bd[Ranking instantiation models] \ \\ Let the notation be as usual and $\AAA^+ = \{a\in\AAA\mid a\not\!\!\rhd\ a\}$. $(R,I)$ is called a ranking instantiation model (more sloppily, a ranking model) of $\Ax$ iff 
\bc
$R\models_{rk}\Delta^{\AAA,I} = \{\ph_a\leadsto\psi_a, \ph_a\LI\theta_a\mid a\in\AAA\}$, 
\ec
and for all $a,b\in\AAA^+$, $a\rhd b$ iff $a\rhd^{R}_I b$.
Let $\Rx^{\Ax}$ be the collection of all the ranking instantiation models of $\Ax$.
\ed

\noindent That is, over the non-loopy arguments, the semantic-based attack relation $\rhd^{R}_I$ specified by $R,I$ has to correspond exactly to the abstract attack relation $\rhd$. The collection of ranking instantiation models is not meant to change if we add or drop attack links between self-reflective and other arguments because the details are absorbed by the impossible joint contexts. If $\Ax$ and $\Ax'$ share the same 1-loops and the same attack structure over the other arguments, $\Rx^{\Ax} = \Rx^{\Ax'}$.  

It also important to observe, and we will come back to this, that each $\Ax = (\AAA,\rhd)$ admits many ranking instantiation models $(R,I)$, obtained by varying the ranking values or the proposition triples associated with the abstract arguments. 

What can we say about classical types of attack? If we focus on the actual semantic content, rebuttal is characterized by incompatible defeasible consequents, and undermining by a defeasible consequent conflicting with an antecedent. In the ranking context, these two types of attacks can be modeled by constraints expressing necessities. The ranking characterizations are as follows. Recall that $\psi_a\fs\ph_a$, $\psi_b\fs\ph_b$.
   \\[0.05in]
$a$ {\bf rebuts} $b$: \ $R(\psi_a\wedge\psi_b) = \infty$, e.g.~if $\psi_a\fs\neg\psi_b$. \\
$a$ {\bf undermines} $b$: \ $R(\psi_a\wedge\ph_b) = \infty$, e.g.~if $\psi_a\fs\neg\ph_b$. 
   \\[0.05in]
In our simple semantic reading, undermining entails rebuttal because $\psi_b\fs\ph_b$. There are four qualitative attack configurations involving two arguments: $\ph_a\wedge\ph_b$ being compatible with neither, one, or both of $\psi_a,\psi_b$. If $a$ asymmetrically undermines $b$, we have $R(\psi_a\wedge\ph_b) = \infty$ and $R(\psi_b\wedge\ph_a)\neq\infty$, hence $R(\ph_a\wedge\ph_b)\neq\infty$. This implies 
$R\models_{rk}\ph_a\wedge\ph_b\wedge\psi_b\LI\neg\psi_a$ and $R\not\models_{rk}\ph_a\wedge\ph_b\leadsto\psi_a$, i.e.~$b\rhd^R_I a$ and $a\not\!\!\rhd^R_I\ b$ according to our attack semantics. It follows that undermining has no obvious ranking semantic justification if we stipulate that the defeasible claim entails the antecedent. Also note that rebuttal is entailed by symmetric and asymmetric attacks.

\section{Ranking extensions}

Ranking instantiation models offer new possibilities to identify reasonable argumentative positions. Let $(R,I)$ be a model of the framework $\Ax = (\AAA,\rhd)$. In the context of $(R,I)$, a minimal requirement for aceptable argument sets $S\subseteq\AAA$ are coherent premises, i.e.~the doxastic possibility of the joint strict contents $\theta_S = \wedge_{a\in S}(\ph_a\wedge\theta_a)$ w.r.t.~$R$, which means $R(\theta_S)\neq\infty$. This excludes self-attacks, but not conflicts within $S$. $S = \emptyset$ is by definition coherent because $\theta_{\emptyset} = \T$. Given that evidence $\ph_a$ should not be rejected without good reasons, the maximally coherent $S\subseteq\AAA$ are of particular interest and constitute suitable background contexts when looking for extensions. Each $E\subseteq S$ then specifies a proposition 
\bc
	$\psi_{S,E} := \theta_S\ \wedge\ \wedge_{a\in E}\psi_a\ \wedge\ \wedge_{a\in \AAA-E}\neg\psi_a$.
\ec
$\psi_{S,E}$ characterizes those worlds verifying the strict content of the $a\in S$ and exactly the defeasible content of the $a\in E$. Because $a\rhd^R_I b$ implies $R(\psi_a\wedge\psi_b) = \infty$, any conflict $a\rhd b$ in $E$ makes $\psi_{S,E}$ impossible. Note however that $R(\psi_{S,E}) = \infty$ may also result from non-binary conflicts involving multiple arguments, or a specific choice of logically dependent $\ph_a,\psi_a$. What are the most reasonable extension candidates $E\subseteq S\subseteq\AAA$ according to $(R,I)$? One idea is to focus on those $E$ which induce the most plausible $\psi_{S,E}$ relative to $\theta_S$ among all their maximal coherent supersets $S$. 

\bd[Ranking extensions]\ Let $(R,I)$ be a ranking instantiation model of $\Ax = (\AAA,\rhd)$. $E\subseteq\AAA$ is called a ranking-extension of $\Ax$ w.r.t.~$(R,I)$ iff there is a maximal coherent $S\subseteq\AAA$ with $E\subseteq S$ and $R(\psi_{S,E}|\theta_S) = 0$. 
\ed

\noindent Observe that if $S = \emptyset$ is the maximally coherent subset of $\AAA$, then $R(\ph_a) = \infty$ for each $a\in\AAA$ and $E = \emptyset$ is the only ranking extension. While the above definition looks rather decent, a cause of concern may be the great diversity of ranking models $(R,I)$ available for any given $\Ax$. Consider for instance $\Ax = (\{p,q,r\},\{(p,q),(q,r)\})$, i.e.~$p\rhd q\rhd r$. $\Ax$ together with a shallow instantiation $I$ then induces ranking constraints described by the conditionals in
\bc
$\Delta^{\Ax,I} = \{\psi_p\wedge\psi_q\leadsto \F$, $\psi_q\wedge\psi_r\leadsto \F$, $\ph_p\wedge\ph_q\leadsto\psi_p$, $\ph_q\wedge\ph_r\leadsto\psi_q,\ph_p\leadsto\psi_p,\ph_q\leadsto\psi_q,\ph_r\leadsto\psi_r\}$. 
\ec
If we assume that each set $\{\ph_x,\psi_x\}$ is logically independent from all the other $\{\ph_y,\psi_y\}$, then $\Delta^{\Ax,I}$ admits a unique justifiably constructible model, which therefore automatically must be the JZ- and JJR-model: $R^{\Ax,I}_{jz}$. In this example it is obtained by minimally shifting the violation areas of the conditionals.
\bc
$R^{\Ax,I}_{jz} = R_0+\infty[\psi_p\wedge\psi_q]+\infty[\psi_q\wedge\psi_r]+1[\ph_p\wedge\ph_q\wedge\neg\psi_p]+ 1[\ph_q\wedge\ph_r\wedge\neg\psi_q]+1[\ph_p\wedge\neg\psi_p]+1[\ph_q\wedge\neg\psi_q]+1[\ph_r\wedge\neg\psi_r]$.
\ec
Given that $S = \AAA$ is coherent, there are eight extension candidates. For the doxastically possible alternatives, $R^{\Ax,I}_{jz}(\psi_{\AAA,\{p,r\}}) = 2 < 3 = R^{\Ax,I}_{jz}(\psi_{\AAA,\{p\}}) = R^{\Ax,I}_{jz}(\psi_{\AAA,\{q\}}) < 4 = R^{\Ax,I}_{jz}(\psi_{\AAA,\{r\}}) < 5 = R^{\Ax,I}_{jz}(\psi_{\AAA,\emptyset}) < \infty$. Because $R^{\Ax,I}_{jz}(\ph_S) = 2$, we get $R^{\Ax,I}_{jz}(\psi_{\AAA,\{p,r\}}|\ph_S)  = 0$. The unique ranking extension is therefore $\{p,r\}$, which is also the standard Dung solution. 

However, without any further constraints on the extension generating ranking instantiation model $(R,I)$, we could pick up as an alternative $R = R^{\Ax,I}_{jz}+\infty[\psi_p\wedge\psi_r\wedge\ph_q]$ such that $R(\psi_p\wedge\psi_r\wedge\ph_q) = \infty$, resp.~an $I$ enforcing $\psi_p\wedge\psi_r\wedge\ph_q\fs F$. In both scenarios, the minima would then become $R(\psi_{\AAA,\{p\}}) = R(\psi_{\AAA,\{q\}}) = 3$, imposing the ranking extensions $\{p\},\{q\}$. Because of $R(\psi_{\AAA,\{p,r\}}) = \infty$, the standard extension $\{p,r\}$ would necessarily be rejected. But this violates a hallmark of argumentation, namely the unconditional support of unattacked arguments, like $p$. This shows that we have to control the choice of ranking instantiation models to implement a reasonable ranking extension semantics.

The idea is now to choose on one hand, as our doxastic background, a well-justified canonical ranking measure model of the default base $\Delta^{\Ax,I}$, e.g.~the JZ-model $R^{\Ax,I}_{jz}$, and on the other hand, implementing Ockham's razor, the simplest instantiations of the given framework $\Ax$. In particular, we stipulate that the instantiations of individual arguments should by default be logically independent. We emphasize that the goal here is just to interpret abstract argumentation frameworks with a minimal amount of additional assumptions, not to adequately model specific real-world arguments. 

We can satisfy these desiderata by using disjoint vocabularies for instantiating different abstract arguments, and by relying on elementary instances of the defeasible modus ponens for the corresponding inference pairs. That is, we introduce for each $a\in\AAA$ independent propositional atoms $X_a,Y_a$, and set $I_{log}(a) = (\{X_a\}\cup\{X_a\leadsto Y_a\},Y_a)$. The corresponding shallow semantic instantiation is then $I(a) = I_{sem}(a) = (\ph_a,\ph_a,\psi_a) = (X_a,X_a,X_a\wedge Y_a)$. We call $I$ a {\it generic instantiation}. Up to boolean isomorphy, it is completely characterized by the cardinality of $\AAA$.  

If we fix a generic instantiation $I$, then the unique justifiably constructible ranking model of $\Delta^{\Ax,I}$ is ($a\lhd\!/\!\rhd b$: $a\rhd b$ or $b\rhd a$)
\bc
$R_{jz}^{\Ax} = R_0+\Sigma_{a\not\rhd a}1[\ph_a\wedge\neg\psi_a]+\Sigma_{a\rhd a}\infty[\ph_a\wedge\neg\psi_a]+\Sigma_{a\rhd b}1[\ph_a\wedge\ph_b\wedge\neg\psi_a]+\Sigma_{a\lhd\!/\!\rhd b}\infty[\psi_a\wedge\psi_b]$ \\
$= R_0+\Sigma_{a\not\rhd a}1[X_a\wedge\neg Y_a]+\Sigma_{a\rhd a}\infty[X_a\wedge\neg Y_a]+\Sigma_{a\rhd b}1[X_a\wedge X_b\wedge\neg Y_a]+\Sigma_{a\lhd\!/\!\rhd b}\infty[X_a\wedge Y_a\wedge X_b\wedge Y_b]$. 
\ec
Because the $\{X_a,Y_a\}$ are logically independent for distinct $a$, and the defaults expressing an attack $a\rhd b$ just concern $X_a\wedge X_b$, only those $X_a$ with $a\rhd a$ become impossible. In fact, $\{\ph_a\leadsto\psi_a,\psi_a\wedge\psi_a\leadsto\F\}\fs_{rk}\ph_a\leadsto\F$. Hence, in line with intuition, the ranking instantiation model $(R_{jz}^{\Ax},I)$ will trivialize exactly the self-defeating arguments. Assuming genericity, $\AAA^+ = \{a\in\AAA\mid a\not\!\!\rhd\ a\}$ is then the unique maximal coherent subset of $\AAA$. We are now ready to specify our ranking-based evaluation semantics. Note that all the generic $I$ are essentially equivalent. \\

\noindent {\bf JZ-evaluation semantics (JZ-extensions):} \\ 
$\Ex_{jz}(\Ax) = \{E\subseteq\AAA\mid E$ ranking extension w.r.t.~$(R^{\Ax,I}_{jz},I)$ for any/all generic $I\}$. \\

\noindent There is actually a simple algorithm to identify the JZ-extensions using extension weights. 

\bd[Extension weight] For each argumentation framework $\Ax = (\AAA,\rhd)$, the extension weight function $r_{\Ax}:2^{\AAA}\I [0,\infty]$ is defined as follows: If $E$ is conflict-free, \\ 
$r_{\Ax}(E) = |\AAA^+ -E| + |\{a\in\AAA^+-E\mid\exists b\in\AAA^+(a\rhd b\wedge \\ \noindent b\not\!\rhd a)\}|$, if not, $r_{\Ax}(E) = \infty$.
\ed

\noindent It is not too difficult to see that $r_{\Ax}(E) = R^{\Ax,I}_{jz}(\psi_{\AAA^+,E})$. Hence, $E\in\Ex_{jz}(\Ax)$ iff $r_{\Ax}(E) = min\{r_{\Ax}(X)\mid X\subseteq\AAA\}$. That is, the JZ-extensions are those where the sum of the number of non-reflective non-extension arguments and the number of one-sided attacks starting from them is minimal.

\section{Examples and properties}

To get a better understanding of the ranking extension semantics and its relation with other extension concepts, let us first take a look at how it handles some basic examples. Because of its uncommon semantic perspective and its partly quantitative character, we will observe some unorthodox behaviour. Under instantiation genericity, it is enough to compare $R^{\Ax,I}(\psi_{\AAA^+,E})$ for $E\subseteq\AAA^+$, or to focus on 1-loop-free  frameworks. For each instance, we specify the domain $\AAA$ and the full attack relation $\rhd$. $\psi_{\AAA^+,\{x_1\ldots x_n\}}$ is abbreviated by $\psi_{x_1,\ldots, x_n}$ resp.~$\psi_{\emptyset}$. 
   \\[0.1in]
\noindent {\bf Simple reinstatement:} \ $\{a,b,c\}$ with $a \rhd b \rhd c$. 
   \\[0.1in]
\noindent The grounded extension $\{a,c\}$ is the canonical result put forward by any standard acceptability semantics. The unique JJ-model, i.e.~the JZ-model $R$ of $\Delta^{\Ax,I}$, satisfies $R(\psi_{a}) = R(\psi_{b}) = 3, R(\psi_{c}) = 4, R(\psi_{a,c}) = 2$, and $R(\psi_{\emptyset}) = 5$. The other candidates all get rank $\infty$. Because $R(\psi_{a,c})$ is minimal, $\{a,c\}$ is the only JZ-ranking extension, i.e.~$\Ex_{jz}(\Ax) = \{\{a,c\}\}$. 
   \\[0.1in]
\noindent {\bf 3-loop:} \ $\{a,b,c\}$ with $a \rhd b \rhd c \rhd a$. 
   \\[0.1in]
\noindent Semantics under the admissibility dogm reject $\{a\},\{b\},\{c\}$, only $\emptyset$ is admissible. But the JZ-model $R$ verifies $R(\psi_{a}) = R(\psi_{b}) = R(\psi_{c}) = 4 < 5 = R(\psi_{\emptyset})$. Because all the alternatives are set to $\infty$, our ranking extensions are the maximal conflict-free sets $\{a\},\{b\},\{c\}$, i.e., $\Ex_{jz}$ clearly violates admissibility. 
   \\[0.1in]
\noindent {\bf Attack on 2-loop:} \ $\{a,b,c\}$ with $a \rhd b \rhd c \rhd b$. 
   \\[0.1in]
\noindent We have $R(\psi_{\emptyset}) = 4, R(\psi_{a}) = 2, R(\psi_{b}) = R(\psi_{c}) = 3, R(\psi_{a,c}) = 1$, but $\infty$ for the others. Here $\Ex_{jz}(\Ax)$ = $\{\{a,c\}\}$ picks up the canonical stable extension. 
   \\[0.1in]
\noindent {\bf Attack from 2-loop:} \ $\{a,b,c\}$ with $b \rhd a \rhd b \rhd c$. 
   \\[0.1in]
\noindent We get $R(\psi_{\emptyset}) = 4, R(\psi_{a}) = 3, R(\psi_{b}) = 2, R(\psi_{c}) = 3$, $R(\psi_{a,b}) = R(\psi_{b,c}) = \infty$, and $R(\psi_{a,c}) = 2$. $\Ex_{jz}(\Ax) = \{\{b\},\{a,c\}\}$ thus collects the stable extensions. 
   \\[0.1in]
\noindent {\bf 3,1-loop:} \ $\{a,b,c\}$ with $a \rhd a \rhd b \rhd c \rhd a$. 
   \\[0.1in]
\noindent $E = \emptyset$ is here the only admissible extension. The maximal coherent set is $\AAA^+ = \{b,c\}$, and we get $R(\psi_{b}) = 1, R(\psi_{c}) = 2$, as well as $R(\psi_{\emptyset}) = 3$. It follows that $\Ex_{jz}(\Ax)=\{\{b\}\}$, rejecting the stage extension $\{c\}$. 
   \\[0.1in]
\noindent {\bf 3,2-loop:} \ $\{a,b,c\}$ with $b \rhd a \rhd b \rhd c \rhd a$.
    \\[0.1in]
\noindent We have $R(\psi_{\emptyset}) = 5, R(\psi_{a}) = 4, R(\psi_{b}) = 3$, and $R(\psi_{c}) = 3$, i.e.~$\Ex_{jz}(\Ax) = \{\{b\},\{c\}\}$. But the stable extension $\{b\}$ is the only admissible ranking extension. 
   \\[0.1in]
\noindent The previous examples show that the ranking extension semantics $\Ex_{jz}$ can diverge considerably from all the other major proposals found in the literature. It may look as if the main difference is its more liberal attitude towards some non-admissible, but still justifiable extensions. However, the semantics has an even more exotic flavour. Consider the following scenarios, where we indicate the minimal extension weights $r_{\Ax}(E)$.
   \\[0.1in]
\noindent {\bf 2-loop chain:} \ $\{a,b,c\}$, $b \rhd a \rhd b \rhd c \rhd b$ : \\ $r(\{a,c\}) = 1 < 2 = r(\{b\})$.                      
    \\[0.1in]
\noindent {\bf Splitted 3-chain:} \ $\{a,b,c,d\}$, $a \rhd b \rhd c$, $a \rhd d \rhd c$ : \\ $r(\{a,c\}) = r(\{b,d\}) = 4$.     
   \\[0.1in]
\noindent {\bf Spoon:} \ $\{a,b,c,d\}$, $a \rhd b \rhd c \rhd d \rhd c$ : \\ $r(\{a,d\}) = r(\{a,c\}) = r(\{b,d\}) = 3$.     
    \\[0.1in]
The first example documents the rejection of a stable extension, namely $\{b\}$. The second one illustrates the impact of quantitative considerations when dealing with a splitted variant of simple reinstatement. The third instance shows the coexistence of two stable extension with a non-admissible one. That is, even attack-free $a$ can be questioned under certain circumstances. It follows that the above ranking semantic interpretation of argumentation frameworks deviates considerably from standard accounts and expectations. Let us now investigate how $\Ex_{jz}$ handles some common principles for extension semantics. 
   \\[0.1in]
\noindent {\bf Isomorphy:} $f:\Ax\cong\Ax'$ \ implies \ $f'':\Ex(\Ax')\cong\Ex(\Ax)$. 
   \\[0.1in]
\noindent {\bf Conflict-freedom:} If $a,b\in E\in\Ex(\Ax)$, then $a\not\!\rhd\ b$. 
   \\[0.1in]
\noindent {\bf CF-maximality:} \ If $E\in\Ex(\Ax)$, then $E$ is a maximal conflict-free subset of $\AAA$.
   \\[0.1in]
\noindent {\bf Inclusion-maximality:} \ If $E,E'\in\Ex(\Ax)$ and $E\subseteq E'$, then $E = E'$. 
    \\[0.1in]
\noindent {\bf Reinstatement:} \ If $E\in\Ex(\Ax)$, $a\in\AAA$, and for each $b\rhd a$ there is an $a'\in E$ with $a'\rhd b$, then $a\in E$. 
   \\[0.1in]
\noindent {\bf Directionality:} \ Let $\Ax_1 = (\AAA_1,\rhd_1),\Ax_2 = (\AAA_2,\rhd_2)$ be such that $\AAA_1\cap\AAA_2 = \emptyset$, $\rhd_0\subseteq A_1\times A_2$, $\Ax = (\AAA_1\cup\AAA_2,\rhd_1\cup\rhd_0\cup\rhd_2)$. Then we have $\Ex(\Ax_1) = \{E\cap\AAA_1\mid E\in\Ex(\Ax)\}$. 

\bt[Basic properties] \ \\
\noindent $\Ex_{jz}$ verifies isomorphy, conflict-freedom, inclusion maximality, and CF-maximality. It falsifies reinstatement and directionality.
\et

\noindent The first four features are easy consequences of the $\Ex_{jz}$-specification. The violation of reinstatement directly follows from how the semantics handles 3-loops. The spoon example documents the failure of directionality if we set $\AAA_1 = \{a,b\}$. But directionality also fails for other prominent approaches, like the semi-stable semantics. Note however that it can be indirectly enforced by using $\Ex_{jz}$ as the base function for an SCC-recursive semantics [BGG 05]. 

The following properties are inspired by the cumulativity principle for nonmonotonic reasoning. They state that if we drop an argument rejected by every extension, then this shouldn't add or erase skeptically supported arguments. 
   \\[0.1in]
\noindent {\bf Rejection cumulativity:} ($\Ax|B$: $\Ax$ restricted to $B$) \\
{\bf Rej-Cut:} If $a\not\in\cup\Ex(\Ax)$, then $\cap\Ex(\Ax|\AAA-\{a\})\subseteq\cap\Ex(\Ax)$. \\
{\bf Rej-CM:} If $a\not\in\cup\Ex(\Ax)$, then $\cap\Ex(\Ax)\subseteq\cap\Ex(\Ax|\AAA-\{a\})$. 
   \\[0.1in]
\noindent Although our semantics relies on default inference notions verifying cumulativity at the level of $\fp^{\Ix}_{\Delta}$, it nevertheless fails to validate the previous postulates.

\bt[No rejection cumulativity]\ \\ 
$\Ex_{jz}$ violates Rej-Cut and Rej-CM.
\et

\noindent The counterexample for Rej-CUT is provided by $b\rhd c\rhd a\rhd b\rhd a$, because $\{b\}\not\subseteq\{b\}\cap\{c\}$. The one for Rej-CM is obtained by adding $c\rhd b$. Here $\{c\}\not\subseteq\{b\}\cap\{c\}$. 

Another idea for combining plausibilistic default reasoning and argumentation theory has been presented in [KIS 11]. It combines defeasible logic programming with a prioritization criterion based on System Z. While it handles some benchmarks better than the individual systems do, its heterogeneous character makes it hard to assess. It doesn't share our goal to seek a plausibilistic semantics for abstract argumentation and seems to produce different results even in the generic context. 

\section{Conclusions}

We have shown how the ranking construction paradigm for default reasoning can be exploited to interpret abstract argumentation frameworks and to specify corresponding extension semantics by using generic argument instantiations and distinguished canonical ranking models. We have considered structured and conditional logical instantiations, defined attack between inference pairs, and after a further abstraction step, introduced simple semantic instantiations, which interpret arguments by triples of premise, strict, and defeasible content. While our basic ranking extension semantics $\Ex_{jz}$ is intuitively appealing and has some interesting properties, it also exhibits a surprisingly unorthodox behaviour. This needs further exploration to see whether there are approaches which share the same semantic spirit but can avoid abnormalities conflicting with the standard argumentation philosophy. Actually, we have been able to develop an alternative semantics which seems to meet these demands, but it will have to be discussed elsewhere.

\section{Bibliography}

\begin{description}

\item[BGG 05] P. Baroni, M. Giacomin, G. Guida. SCC-recursiveness: a general schema for argumentation semantics. AIJ 168:163-210, 2005.
\item[BSS 00] S. Benferhat, A. Saffiotti, P. Smets. Belief functions and default reasoning. Artificial Intelligence 122(1-2): 1-69, 2000.
\item[GMP 93] M. Goldszmidt, P. Morris, J. Pearl. A maximum entropy approach to nonmonotonic reasoning. IEEE Transact. Patt. Anal. and Mach. Int, 15:220-232, 1993.
\item[KI 01] G. Kern-Isberner. Conditionals in nonmonotonoic reasoning and belief revision, LNAI 2087. Springer, 2001.
\item[KIS 11] G. Kern-Isberner G.R. Simari. A Default Logical Semantics for Defeasible Argumentation. Proc.~of FLAIRS 2011, AAAI Press, 2011.
\item[KLM 90]  S. Kraus, D. Lehmann, M. Magidor. Nonmonotonic reasoning, preferential models and cumulative logics. In {\it Artificial Intelligence}, 44:167-207, 1990.
\item[Mak 94] D. Makinson. General patterns of nonmonotonic reasoning. Handbook of Logic in AI and LP, vol. 3 (eds. Gabbay et al.): 35-110. Oxford University Press, 1994.
\item[Pea 90] J. Pearl. System Z: a natural ordering of defaults with tractable applications to nonmonotonic reasoning. TARK 3: 121-135. Morgan Kaufmann, 1990.
\item[Spo 88] W. Spohn. Ordinal conditional functions: a dynamic theory of epistemic states. Causation in Decision, Belief Change, and Statistics (eds. W.L. Harper, B. Skyrms): 105-134. Kluwer, 1988.
\item[Spo 12] W. Spohn. The Laws of Belief. Ranking Theory and Its Philosophical Applications. Oxford University Press, Oxford 2012.
\item[Wey 94] E. Weydert. General belief measures. UAI'94, Morgan Kaufmann.
\item[Wey 95] E. Weydert. Defaults and infinitesimals. Defeasible inference by non-archimdean entropy maximization. UAI 95: 540-547. Morgan Kaufmann, 1995.
\item[Wey 96] E. Weydert. System J - rev. entailment. FAPR 96:637-649. Springer, 1996.
\item[Wey 98] E. Weydert. System JZ - How to build a canonical ranking model of a default knowledge base. KR 98: 190-201. Morgan Kaufmann, 1998.
\item[Wey 03] E. Weydert. System JLZ - Rational default reasoning by minimal ranking constructions. Journal of Applied Logic 1(3-4): 273-308. Elsevier, 2003.  
\item[Wey 11] E. Weydert. Semi-stable extensions for infinite frameworks. In Proc. BNAIC 2012: 336–343.
\item[Wey 13] E. Weydert. On the Plausibility of Abstract Arguments. ECSQARU 2013, LNAI 7958 (ed. L. van der Gaag): 522-533 Springer, 2013.
\end{description}

\end{document}